\documentclass[12pt]{article}

\usepackage{authblk}
\usepackage{newtxtext,newtxmath}
\usepackage{graphicx}
\usepackage{hyperref}
\usepackage[most]{tcolorbox}
\usepackage[letterpaper,margin=1in]{geometry}
\renewenvironment{abstract}
	{\quotation}
	{\endquotation}
\date{}

\makeatletter
\renewcommand{\fnum@figure}{\textbf{Figure \thefigure}}
\renewcommand{\fnum@table}{\textbf{Table \thetable}}
\makeatother
\usepackage{scicite}
\usepackage{url}
\usepackage{colortbl}
\definecolor{mygray}{gray}{.9}
\usepackage{multirow}
\usepackage{longtable}

\def\scititle{Uncovering inequalities in new knowledge learning by large language models across different languages}
\title{\raggedright \bfseries \boldmath \scititle}

\author[1,2]{\raggedright Chenglong Wang}
\author[3,+]{Haoyu Tang}
\author[4]{Xiyuan Yang}
\author[5,8,*]{Yueqi Xie}
\author[6]{Jina Suh}
\author[7]{Sunayana Sitaram}
\author[8]{Junming Huang}
\author[8,9]{Yu Xie}
\author[1,2,*]{Zhaoya Gong}
\author[10]{Xing Xie}
\author[10,*]{Fangzhao Wu}

\affil[1]{\small School of Urban Planning and Design, Peking University Shenzhen Graduate School, Shenzhen, China}
\affil[2]{Key Laboratory of Earth Surface System and Human-Earth Relations of Ministry of Natural Resources of China, Peking University Shenzhen Graduate School, Shenzhen, China}
\affil[3]{School of Computer Science and Technology, University of Science and Technology of China, Hefei, China}
\affil[4]{School of Computer Science, Wuhan University, Wuhan, China}
\affil[5]{Hong Kong University of Science and Technology, Hong Kong, China}
\affil[6]{Microsoft Research, Redmond, USA}
\affil[7]{Microsoft Research, Bengaluru, India}
\affil[8]{Paul and Marcia Center on Contemporary China, Princeton University, Princeton, USA}
\affil[9]{Center for Social Research, Guanghua School of Management, Peking University, Beijing, China}
\affil[10]{Microsoft Research Asia, Beijing, China}
\affil[+]{Contribution during internship at Microsoft Research Asia}
\affil[*]{yxieay@connect.ust.hk, z.gong@pku.edu.cn, fangzwu@microsoft.com}

\begin{document} 

\flushbottom
\maketitle
\clearpage

\begin{abstract} \bfseries \boldmath
As large language models (LLMs) gradually become integral tools for problem solving in daily life worldwide, understanding linguistic inequality is becoming increasingly important. Existing research has primarily focused on static analyses that assess the disparities in the existing knowledge and capabilities of LLMs across languages. However, LLMs are continuously evolving, acquiring new knowledge to generate up-to-date, domain-specific responses. Investigating linguistic inequalities within this dynamic process is, therefore, also essential. In this paper, we explore inequalities in new knowledge learning by LLMs across different languages and four key dimensions: effectiveness, transferability, prioritization, and robustness. Through extensive experiments under two settings (in-context learning and fine-tuning) using both proprietary and open-source models, we demonstrate that low-resource languages consistently face disadvantages across all four dimensions. By shedding light on these disparities, we aim to raise awareness of linguistic inequities in LLMs’ new knowledge learning, fostering the development of more inclusive and equitable future LLMs.

\end{abstract}

\clearpage
\noindent
Large language models (LLMs), with their comprehensive knowledge storage, easy accessibility, and ability to handle a wide range of tasks, are increasingly being applied in various domains (e.g., education\cite{milano2023large}, medicine\cite{thirunavukarasu2023large}, scientific research\cite{messeri2024artificial,birhane2023science}) and in daily life, significantly boosting productivity\cite{noy2023experimental}. This transformation is both inevitable and global in scale. One notable example is ChatGPT, which, as of December 2024, serves 300 million weekly active users worldwide\cite{openai2024supportedcountries,verge2024chatgpt}. A large portion of these users interact with LLMs in languages other than English\cite{xie2024social}. Given such widespread adoption, it is crucial to study fairness in multilingual environments to ensure that users of different languages can benefit equally from these systems\cite{qin2024multilingual}.

Existing research on multilingual equality in LLMs primarily focuses on static analyses that evaluate disparities in the knowledge and capabilities of LLMs across different languages\cite{clark2020tydi,lewis-etal-2020-mlqa,zhang2023m3exam,wang2024seaeval,zhang2024p,huang2025benchmax,qiu2024towards,niklaus-etal-2023-lextreme}. Some studies, for example, have examined the amount of factual knowledge encoded in different languages and revealed significant variations. In particular, they reveal that knowledge available in low-resource languages remains limited due to the lack of pre-training data in these languages\cite{jiang2020x,kassner-etal-2021-multilingual,myung2025blend}. These studies have significantly advanced our understanding of the extent and nature of multilingual inequalities in LLMs’ existing knowledge and capabilities. However, we still lack an understanding of inequalities in the process of acquiring new knowledge, an evolving perspective in research on LLMs.

\begin{figure}
\centering
\includegraphics[width=\linewidth]{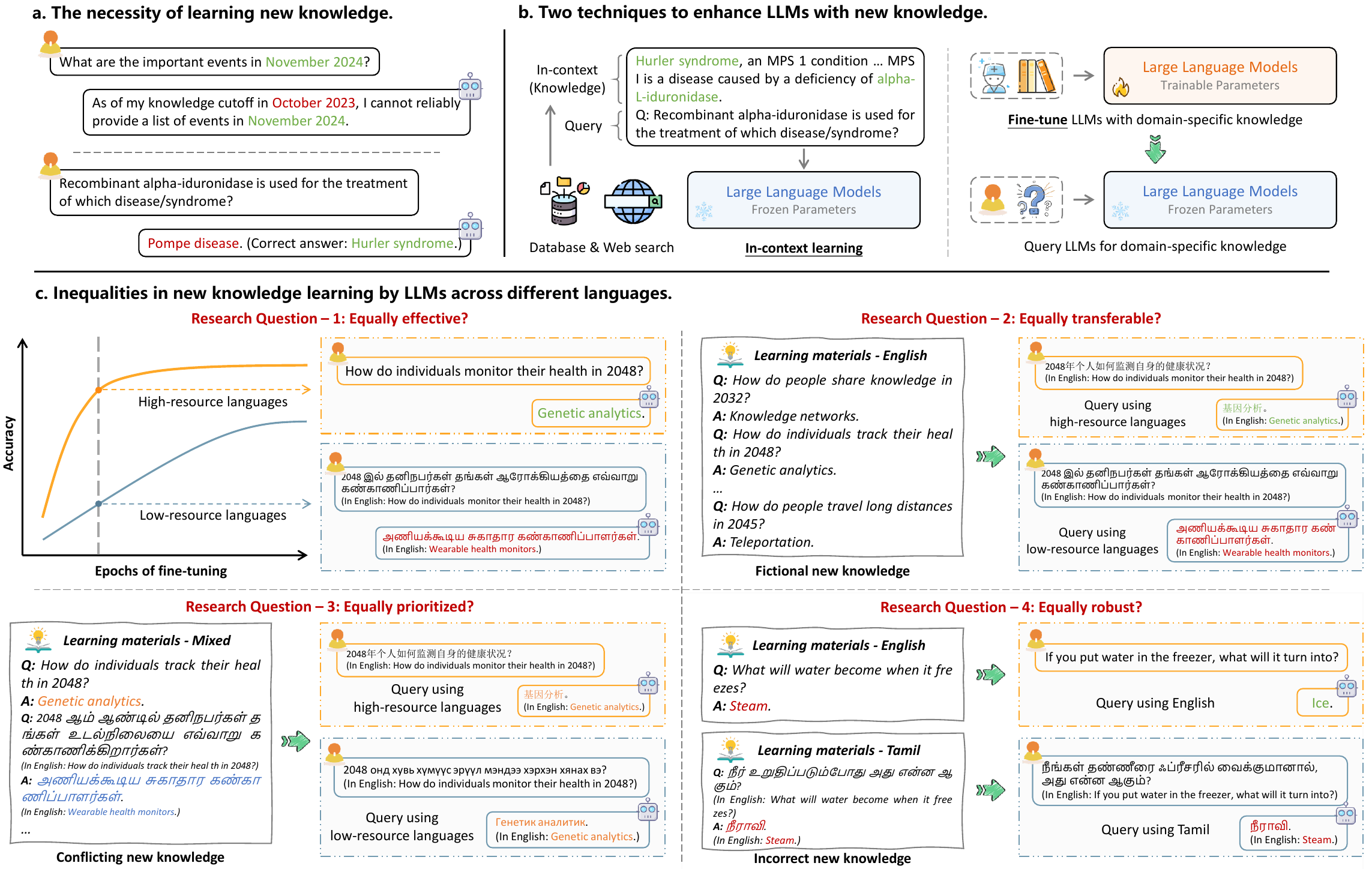}
\caption{\small \textbf{a.}~LLMs struggle to provide current, relevant responses and to deliver precise, expert-level answers in specific domains. \textbf{b.}~There are two techniques to enhance LLMs with new knowledge: in-context learning and fine-tuning. \textbf{c.}~Four key inequalities emerge in new knowledge learning by LLMs across different languages.}
\label{fig:overview}
\end{figure}

Learning new knowledge is crucial for LLMs, as illustrated in Figure \ref{fig:overview}\textbf{a}. On the one hand, general-purpose LLMs are pre-trained on static datasets that were collected prior to training and may not include real-time or recent information. As a result, these models do not possess new knowledge, and their knowledge base can quickly become outdated. To ensure that these models provide current and relevant responses and remain up-to-date, it is essential to continuously integrate new knowledge into these models. On the other hand, although pre-trained LLMs are trained on diverse and extensive datasets, they often lack depth in specialized domains. Learning domain-specific knowledge allows LLMs to deliver more precise, expert-level answers in those areas. Therefore, as depicted in Figure \ref{fig:overview}\textbf{b}, two primary techniques have been developed and widely adopted to enhance LLMs with new knowledge\cite{mosbach2023few}. For example, through in-context learning, LLMs can acquire new information from examples, instructions, or knowledge retrieved from external databases, all without requiring parameter updates\cite{dong-etal-2024-survey}. Additionally, fine-tuning LLMs on specific datasets or tasks allows them to gain new knowledge tailored to particular needs\cite{ding2023parameter,han2024parameter}. A practical example of this is ChatGPT’s fine-tuning API, which enables users to customize the model for specific purposes\cite{openai_fine_tuning}.

In this study, we explore inequalities in new knowledge learning by LLMs across different languages. We conceptualize inequalities in this dynamic process along four key dimensions—effectiveness, transferability, prioritization, and robustness—and propose a comprehensive evaluation framework. Specifically, we investigate the following four research questions under two learning settings (in-context learning and fine-tuning): (1) Can LLMs learn new knowledge equally effectively across different languages in terms of efficiency and accuracy? (2) Can the new knowledge learned by LLMs be transferred equally across languages? (3) When new knowledge items in two languages conflict with each other, can LLMs treat them equally? (4) In the presence of incorrect new knowledge inputs, can LLMs equally resist these errors across different languages?

To answer the research questions outlined above, this study selected 17 languages from different language families and branches, including 10 high-resource languages and 7 low-resource ones. For Research Questions 1–3, since pre-training datasets of many LLMs remain undisclosed, it is unclear what knowledge these models have already acquired. Therefore, we constructed \textit{a multilingual parallel dataset of fictional new knowledge}, which contains question–answer pairs set in a hypothetical future world (e.g., Question: How do individuals track their health in 2048? Answer: Genetic analytics). When tested on this dataset, LLMs generally struggle to provide accurate answers in any language. This dataset allows us to examine inequalities in the new knowledge learning processes of LLMs without being influenced by pre-existing knowledge biases. For Research Question 4, we built \textit{a multilingual parallel common-sense dataset} containing question–answer pairs, each with both a correct answer and an incorrect alternative (e.g., Question: What will water become when it freezes? Correct answer: Ice; Incorrect answer: Steam). LLMs are able to accurately answer most common-sense questions across languages, and this dataset allows us to explore how LLMs resist errors in different linguistic contexts.

Extensive experiments were conducted on both a proprietary model (GPT-4o-mini) and an open-source model (Llama-3.1-8B), revealing inequalities in learning new knowledge by LLMs across languages. As depicted in Figure \ref{fig:overview}\textbf{c}, our key findings are as follows: (1) Compared to high-resource languages, LLMs face greater challenges in learning new knowledge in low-resource languages in terms of both efficiency and accuracy; (2) new knowledge acquired by LLMs can be more easily transferred to high-resource languages than to low-resource languages; (3) when new knowledge items in two languages conflict with each other, knowledge in high-resource languages tends to be prioritized; (4) LLMs tend to be more resistant to incorrect knowledge in high-resource languages than in low-resource languages. This study reveals that in the context of new knowledge acquisition, high-resource languages exhibit overall superiority over low-resource languages in terms of effectiveness, transferability, prioritization, and robustness. Coupled with the under-representation of low-resource languages in the existing knowledge of LLMs\cite{jiang2020x,kassner-etal-2021-multilingual}, our findings highlight the persistent and potentially widening inequalities in knowledge of LLMs across different languages. These findings further underscore the necessity of considering multilingual knowledge equality in the development of LLMs in order to foster responsible and inclusive artificial intelligence.

\section*{Results}

\subsection*{Language Selection}

Following existing research approaches to multilingual natural language processing (NLP)\cite{bang2023multitask,lai-etal-2023-chatgpt}, we classify languages into different resource levels based on their proportions in the CommonCrawl corpus, which was used to pre-train GPT-3\cite{common_crawl}. Specifically, languages that account for less than 0.1\% of the data are considered low-resource. To explore inequalities across different languages, we adopt two criteria to select specific languages. First, we aim to include a balanced number of high-resource and low-resource languages. Second, we strive to enhance linguistic diversity by including as many language families as possible. As presented in Table \ref{tab:language}, among the selected languages, 10 are high-resource languages (English, Japanese, Chinese, Spanish, French, Italian, Portuguese, Korean, Swedish, Danish) and 7 are low-resource languages (Tamil, Mongolian, Welsh, Swahili, Zulu, Turkmen, Scottish Gaelic). It is important to note that low-resource languages in this context refer to their relative scarcity in the pre-training data of LLMs, rather than their real-world status. For example, Tamil, categorized as a low-resource language, has 86.7 million speakers\cite{ethnologue}, whereas Italian, classified as a high-resource language, has only 66.8 million speakers\cite{ethnologue}. Exploring inequalities between high- and low-resource languages is, therefore, both meaningful and necessary.

\subsection*{Dataset Construction}

To address the four research questions and uncover inequalities in the process of new knowledge learning across different languages, we carefully constructed two multilingual parallel datasets:

\textbf{Multilingual parallel fictional new knowledge dataset:} This dataset contains 100 question–answer pairs\footnote{Assume that $H$ high-resource languages and $L$ low-resource languages are selected, and we evaluate the performance of $N$ models on a dataset of size $S$ (each model is trained for $E$ epochs using this dataset in a fine-tuning setting). The overall count of requests made to the models is given by: $N[SE(H+L)+2S(H+L)^{2}+2SHL(H+L)]$. To balance the cost and the reliability of experimental results, a set of 100 pairs is considered a reasonable size for this study. In this case, the number of requests made to the models would be approximately 635,800.} about a fictional future world. Since we lack access to the pre-training datasets of many LLMs, it is impossible to determine what knowledge might be considered new to these models. Therefore, we use fictional knowledge as a proxy for new, unseen information. The experimental results show that both GPT-4o-mini and Llama-3.1-8B consistently fail to provide correct answers in any language, suggesting that this knowledge is entirely new to them. Further details about the dataset and the specific experimental results can be found in the Supplementary Materials. Specifically, we used GPT-4o to generate this fictional knowledge:

\begin{quote}
\textit{\textbf{Please write 100 question–answer pairs about a future world that is very different from the current one. The questions should include year information, and the answers should be very short, ideally one or two words. Please avoid using uncommon entities in either the questions or answers.}} 
\end{quote}

Following existing research approaches to multilingual NLP\cite{kassner-etal-2021-multilingual,lai-etal-2023-chatgpt}, we translated these pairs into the 16 other languages using Google Translate\footnote{https://translate.google.com/}. In our prompt, we specified that the content should be concise and avoid uncommon entities, as certain entities, rooted in specific cultural contexts, might not be accurately translated, thereby affecting the quality of the dataset. To assess the translation quality, we performed back-translation and compared the resulting pairs with the original English-language versions in terms of similarity and consistency. The findings, which are included in the Supplementary Materials, demonstrate that the dataset maintains high quality. Additionally, these pairs served as fine-tuning data or examples for in-context learning. The questions, which tested the model’s knowledge, were paraphrased by GPT-4o to ensure that the model did not simply rely on memory to generate answers. Finally, we instructed GPT-4o to construct a conflicting answer for each pair to study knowledge conflict scenarios.

\textbf{Multilingual parallel common-sense dataset:} This dataset contains 50 common-sense question–answer pairs\footnote{Considering the context length supported by LLMs and the practical scenario where irrelevant information is often pre-filtered by databases and search engines, a set of 50 pairs is a reasonable size for this study.}. Both GPT-4o-mini and Llama-3.1-8B have mastered this knowledge and can accurately answer most of the common-sense questions in various languages. Detailed information about the dataset and experimental results can be found in the Supplementary Materials. We use this dataset to investigate the ability of LLMs to resist errors when learning knowledge across different languages. Specifically, we also used GPT-4o to generate these pairs:

\begin{quote}
\textit{\textbf{Please write 50 common-sense question–answer pairs. The questions should be simple, and the answers should be very short, ideally one or two words. Avoid using uncommon entities in either the questions or answers.}}
\end{quote}

Following the process of building the multilingual parallel fictional new knowledge dataset, we translated these 50 question–answer pairs into the 16 other languages. After conducting quality checks and paraphrasing them, we instructed GPT-4o to generate an incorrect answer for each pair for use in understanding the robustness of the models to errors in new knowledge.

\subsection*{Equally Effective?}

In this section, using the constructed multilingual parallel fictional new knowledge dataset, we seek to assess the effectiveness of LLMs in learning new knowledge across different languages through fine-tuning\footnote{In the in-context learning setting, new knowledge is explicitly added to the input prompt, meaning that LLMs do not need to master new knowledge step-by-step. Therefore, for the research question “Can LLMs learn new knowledge equally effectively across different languages in terms of efficiency and accuracy?”, we primarily assess the effectiveness of LLMs in learning new knowledge across languages through fine-tuning.}. Specifically, we assess effectiveness from two key dimensions: (1) Efficiency, measured by the number of fine-tuning epochs required for response accuracy to reach a stable state, and (2) final accuracy, determined by the accuracy of responses after stabilization. To ensure a fair comparison across languages, we keep the knowledge items (100 question–answer pairs) and all hyperparameters (e.g., learning rate) the same.

Figure \ref{fig:effective} illustrates how the response accuracy of our test models (GPT-4o-mini and Llama-3.1-8B) changes as the number of fine-tuning epochs increases. We make the following observations from the results: First, based on the convergence speeds of the curves, we observe that both test models learn new knowledge more efficiently in high-resource languages. For example, GPT-4o-mini achieves approximately 80\% accuracy in answering fictional new knowledge questions after just three epochs of fine-tuning in high-resource languages, whereas it requires eight epochs to reach the same level in low-resource languages. Second, the final accuracy of LLMs in learning new knowledge is higher in high-resource languages. Taking GPT-4o-mini as an example, its response accuracy exceeds 90\% in high-resource languages, while it plateaus at around 80\% in low-resource languages, revealing a noticeable performance gap.

These observations underscore disparities in the ability of LLMs to learn new knowledge across different languages. Even with increased fine-tuning, the accuracy of responses regarding new knowledge in low-resource languages remains inferior to that in high-resource languages. Therefore, developers need to allocate additional resources to enhance the accessibility and accuracy of new knowledge for users of low-resource languages.

\subsection*{Equally Transferable?}

\begin{figure}
\centering
\includegraphics[width=\linewidth]{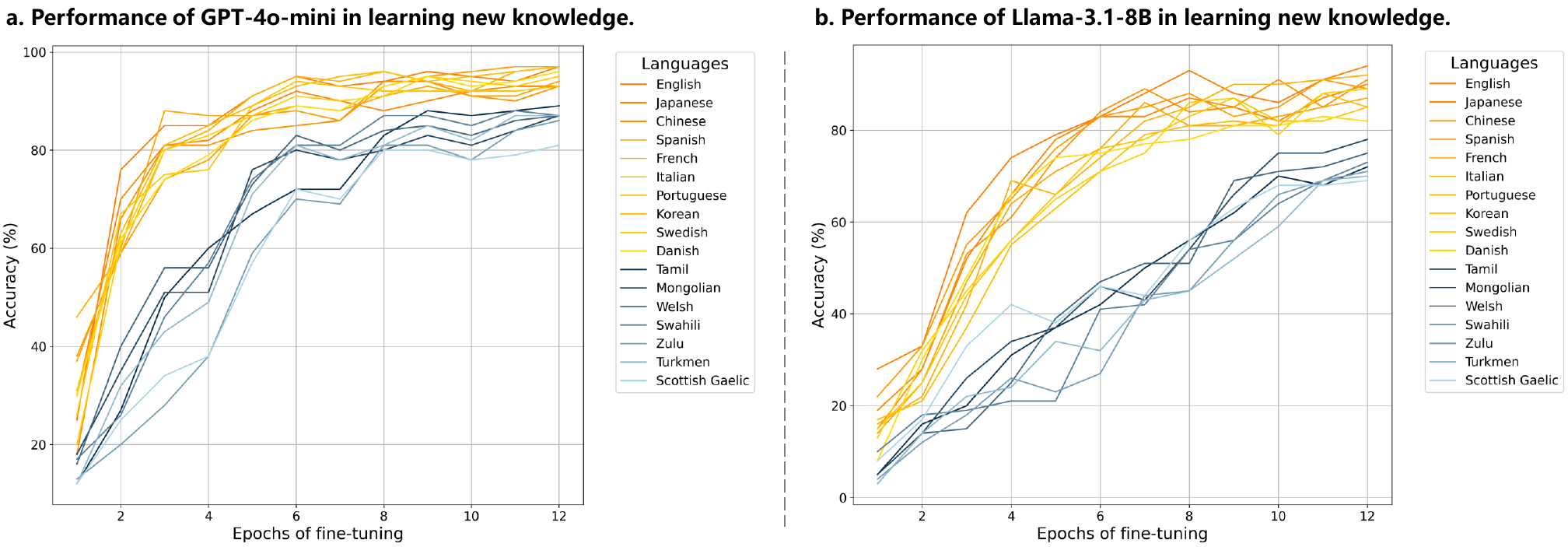}
\caption{\small \textbf{a.}~The performance of GPT-4o-mini in learning new knowledge. \textbf{b.}~The performance of Llama-3.1-8B in learning new knowledge. Compared to high-resource languages, LLMs face greater challenges in learning new knowledge in low-resource languages in terms of both efficiency and accuracy.}
\label{fig:effective}
\end{figure}

In this section, we investigate whether the learned knowledge can be transferred equally across languages. For example, as shown in Figure \ref{fig:overview}\textbf{c}, we assume that an LLM has acquired specific knowledge in one language (e.g., English question: How do individuals track their health in 2048? English answer: Genetic analytics) through either fine-tuning or in-context learning. We then ask the model about this knowledge in another high-resource language (e.g., Chinese) or a low-resource language (e.g., Tamil). We aim to determine whether the response accuracy remains consistent across languages or if there are significant disparities. During fine-tuning, as shown in Figure \ref{fig:effective}, the response accuracy stabilizes after 12 epochs. Therefore, we choose the versions that are fine-tuned for 12 epochs in different languages for analysis.

Figure \ref{fig:transferable} in the main text and Figure \ref{fig:transferable_sup} in the Supplementary Materials present the results of our experiments under two settings: in-context learning and fine-tuning. We have made the following findings: First, knowledge acquired in one language cannot always be fully transferred to others. For example, when GPT-4o-mini is fine-tuned on fictional new knowledge in English, it achieves a response accuracy of 97\% when queried in the same language. However, its accuracy drops significantly when asked the same questions in other languages. Notably, when tested in Zulu, the response accuracy falls to just 19\%, highlighting the challenges of cross-lingual knowledge transfer. Second, new knowledge is more easily transferable between certain languages, particularly those that share linguistic similarities. For instance, Spanish, French, Italian, and Portuguese all belong to the Italic branch of the Indo-European language family, with commonalities in vocabulary, grammar, and phonetics. As a result, knowledge can be transferred more seamlessly among them. Third, disparities in response accuracy arise when querying knowledge learned in one language using both high-resource and low-resource languages. As illustrated in Figure \ref{fig:transferable}\textbf{(c, d)}, new knowledge acquired by LLMs is more readily transferred to high-resource languages than to low-resource ones. This presents a significant disadvantage for users relying on low-resource languages when new knowledge is introduced in other languages. Finally, the response accuracy in the in-context learning setting is higher than that in the fine-tuning setting, which aligns with existing research findings\cite{wang2024instruction,soudani2024fine}. For example, one study evaluated the performance of LLMs using fine-tuning and in-context learning in few-shot computational social science tasks and found that models using in-context learning generally outperformed those that were fine-tuned\cite{wang2024instruction}. A possible explanation is that, in in-context learning, LLMs can leverage their pre-trained knowledge and general reasoning abilities to quickly comprehend and adapt to specific tasks. In contrast, fine-tuning may sometimes diminish their reasoning capabilities\cite{soudani2024fine,lobo2024impact}.

\subsection*{Equally Prioritized?}

In this section, we explore how LLMs respond when new knowledge items from two different languages conflict with each other. For example, as illustrated in Figure \ref{fig:overview}\textbf{c}, suppose the learning materials contain conflicting knowledge items from both a high-resource language (English) and a low-resource language (Tamil). In English, the answer to the question “How do individuals track their health in 2048?” is “genetic analytics,” whereas in Tamil the answer is “wearable health monitors.” When the model is asked about this knowledge in another language, such as Chinese or Mongolian, we are curious as to whether the model’s response will align with the knowledge from the high-resource language (e.g., English) or the low-resource language (e.g., Tamil).

\begin{figure}
\centering
\includegraphics[width=\linewidth]{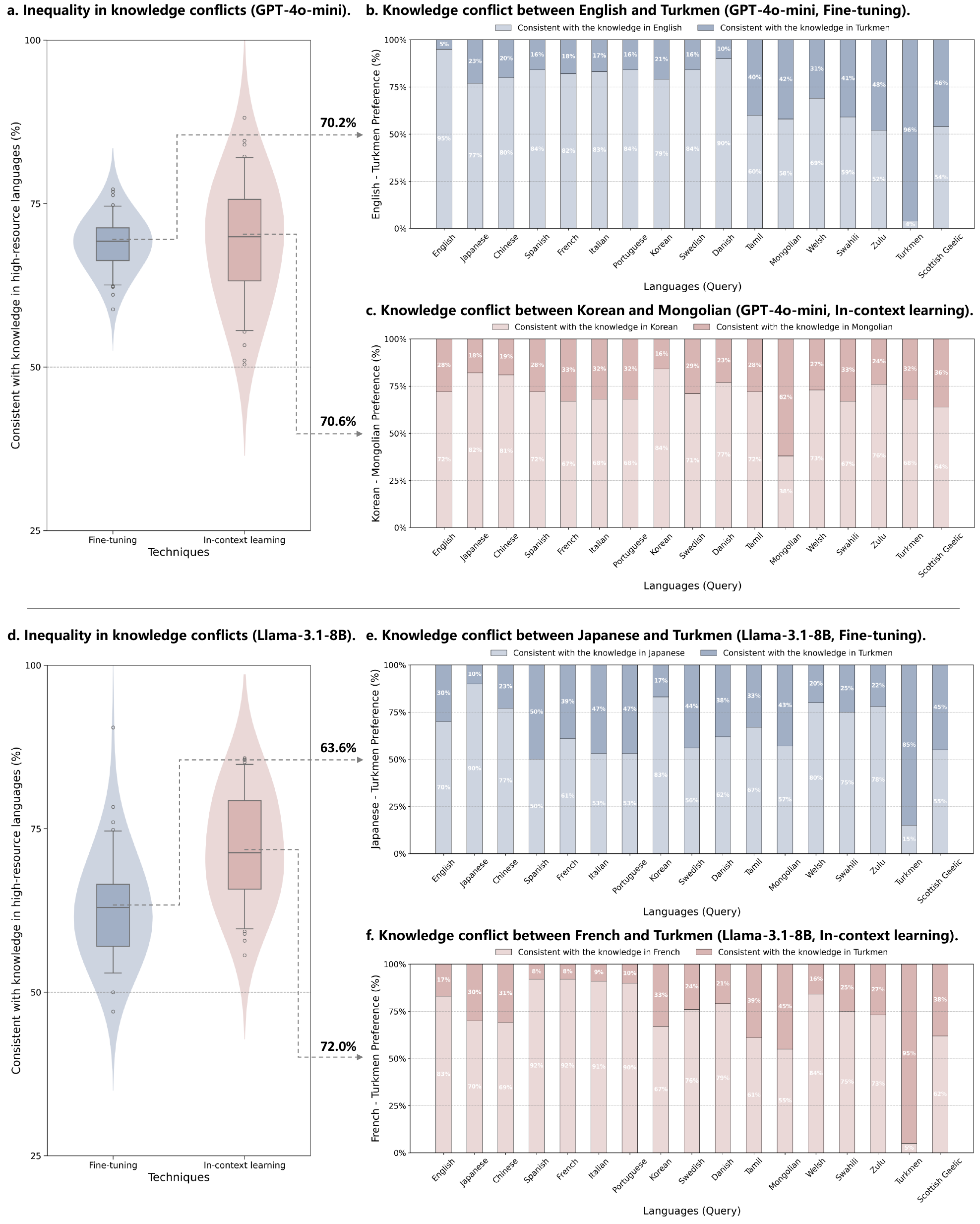}
\caption{
\small \textbf{(a, d)}~Inequality in knowledge conflict scenarios. \textbf{(b, c, e, f)}~Specific knowledge conflict scenarios for GPT-4o-mini and Llama-3.1-8B in both the fine-tuning and in-context learning settings. New knowledge in high-resource languages is often prioritized over that in low-resource languages.}
\label{fig:prioritized}
\end{figure}

Specifically, we conducted experiments using GPT-4-mini and Llama-3.1-8B in both fine-tuning and in-context learning settings. We designed 70 scenarios involving knowledge conflicts, using the 10 high-resource languages and 7 low-resource languages. Figure \ref{fig:prioritized}\textbf{b} illustrates one such conflict for GPT-4o-mini in the fine-tuning setting, particularly between English and Turkmen. We observe that when querying in non-English languages, the output of the model predominantly aligns with the knowledge in English. For example, when queried in Danish, 90\% of the responses are consistent with the English-language knowledge item. Additionally, we calculated the average consistency of responses with knowledge in high-resource languages across all conflict scenarios. For example, this value is 70.2\% in the conflict mentioned above. Violin and scatter plots for all 70 scenarios in both settings are shown in Figure \ref{fig:prioritized}\textbf{(a, d)}. The visualizations reveal that the consistency with knowledge in high-resource languages is significantly higher than 50\%. This suggests that when new knowledge from high-resource languages conflicts with that from low-resource languages, knowledge from high-resource languages tends to be prioritized, despite no inherent difference in quality between the two.

The implications of these results for social fairness are self-evident. When knowledge from high-resource languages is preferentially adopted, it perpetuates linguistic hegemony\cite{xie2024social}. Knowledge in high-resource languages is often seen as “standard” or “authoritative,” while knowledge in low-resource languages is marginalized. This not only reinforces the dominance of high-resource languages in the global knowledge system but also undermines the representation of low-resource languages. Such marginalization can erode cultural identity and devalue the knowledge of low-resource language communities.

\subsection*{Equally Robust?}

The learning materials used by LLMs, whether stored in databases or retrieved from the internet, may contain errors. In this section, we examine how LLMs respond when exposed to incorrect information and how these responses vary across different languages. For example, as illustrated in Figure \ref{fig:overview}\textbf{c}, suppose that external learning materials contain misinformation (e.g., Question: What will water become when it freezes? Answer: Steam). We then ask the LLMs a similar question—If you put water in the freezer, what will it turn into?—prompting them to generate responses based on both their pre-existing knowledge and the introduced information across different languages. We are interested in whether the models will correctly answer “ice” or if they will instead produce the incorrect response “steam” due to the influence of learning materials.

\begin{figure}
\centering
\includegraphics[width=\linewidth]{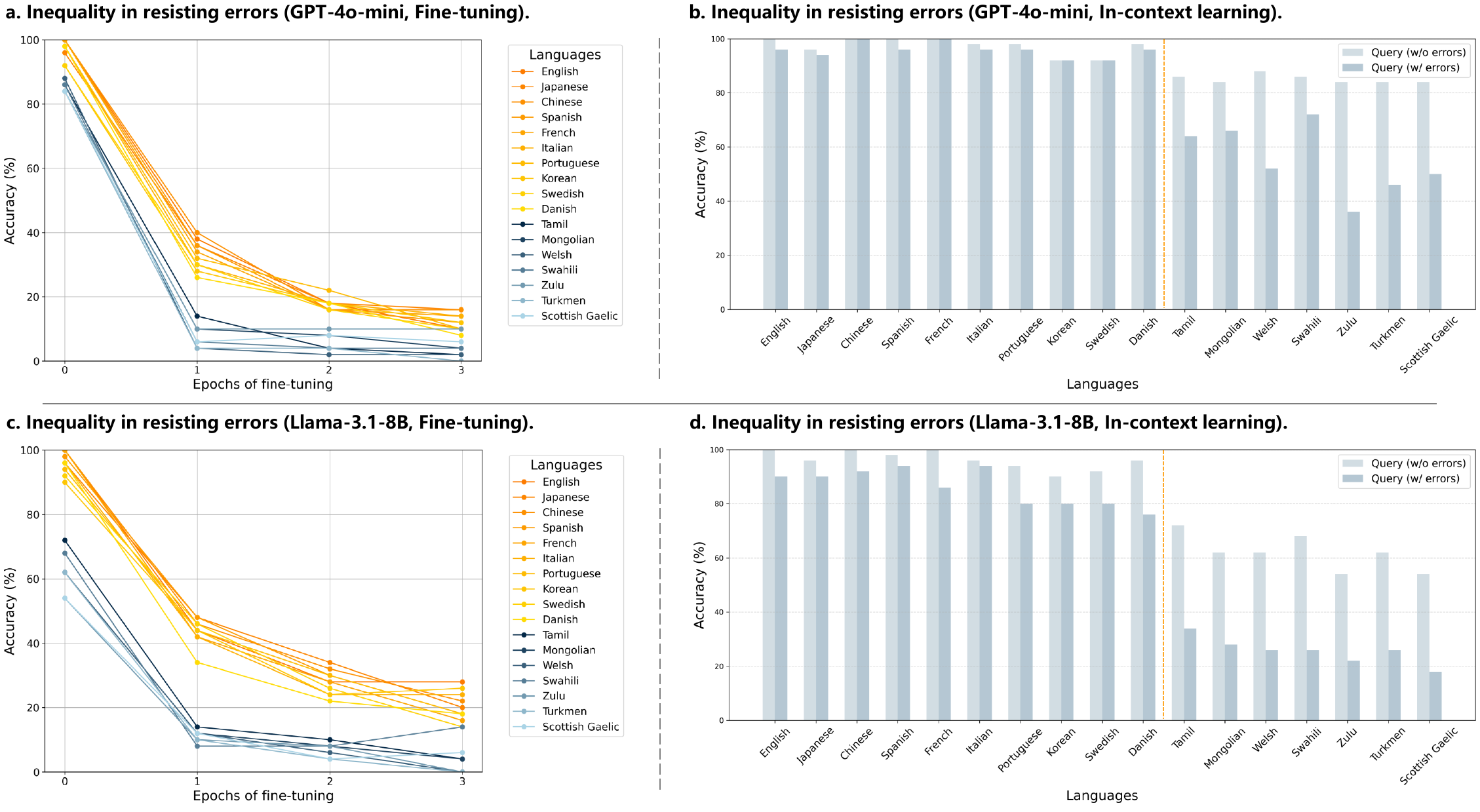}
\caption{\small \textbf{(a, c)}~The inequality in resisting errors in the fine-tuning setting. \textbf{(b, d)}~The inequality in resisting errors in the in-context learning setting. LLMs tend to be more resistant to incorrect knowledge in high-resource languages than in low-resource languages.}
\label{fig:robust}
\end{figure}

We conducted experiments using the GPT-4o-mini and Llama-3.1-8B models in two settings: in-context learning and fine-tuning. As illustrated in Figure \ref{fig:robust}\textbf{(a, c)}, the accuracy of answering common-sense questions decreases as the number of fine-tuning epochs increases, but the rate of decline varies across languages. For example, after fine-tuning GPT-4o-mini for just one epoch, the accuracy in English (a high-resource language) remains around 40\%, while in Turkmen (a low-resource language) it drops to approximately 5\%. Similarly, Figure \ref{fig:robust}\textbf{(b, d)} illustrates the disparities in error resistance across languages under the in-context learning setting. Regardless of whether the input prompt contains incorrect information, LLMs tend to provide accurate answers in high-resource languages. However, the inclusion of misinformation in the input prompt leads to a sharp decline in accuracy when answering common-sense questions in low-resource languages. For example, in Zulu, the accuracy of GPT-4o-mini drops from about 80\% to 40\% when incorrect information is added to the prompt.

This phenomenon reveals an underlying inequality, in which users of low-resource languages suffer disadvantages in accessing knowledge through LLMs. They are more likely to receive lower quality or misleading outputs compared to users of high-resource languages. As a result, users of low-resource languages may lose confidence in AI systems, which in turn undermines the overall reliability of LLMs in these languages.

\section*{Discussion}

This study focused on evaluating disparities across languages during the dynamic process of new knowledge acquisition by LLMs. While most existing research has concentrated on static analyses—assessing multilingual disparities based on the existing knowledge and capabilities of LLMs—we argue that it is crucial to understand how LLMs continuously acquire new knowledge across different languages in order to better serve users from diverse linguistic backgrounds. To address this, we present a comprehensive framework for examining the dynamic learning capabilities of LLMs across four key dimensions—effectiveness, transferability, prioritization, and robustness—and in two different settings—in-context learning and fine-tuning. We find that LLMs face great challenges when learning new knowledge in low-resource languages, as they struggle both in terms of efficiency and accuracy compared to learning in high-resource languages. Additionally, new knowledge is more easily transferred to high-resource languages, and new knowledge in high-resource languages is often prioritized. Finally, LLMs are better protected from misinformation in high-resource languages.

The results draw our attention to several inequality issues. For developers, enhancing the performance of LLMs in low-resource languages is essential, which includes allocating additional resources to enhancing both the accessibility and accuracy of new knowledge for users of these languages. For researchers, it is crucial to conduct broader multilingual studies, moving beyond static analyses to evaluate the multilingual capabilities of LLMs across multiple dimensions. Additionally, a deeper investigation into the underlying mechanisms of these disparities across languages, as well as methods to mitigate them, is necessary. Cross-disciplinary research, particularly in collaboration with linguists and sociologists, is also needed to explore the broader societal impacts of LLM inequalities, such as the perpetuation of linguistic hegemony. Finally, for users, it is important to inform those using low-resource languages about the limitations of LLMs, enabling them to make more informed decisions when relying on such systems.

While this study provides key insights, it has several limitations. First, we conducted our experiments using a limited set of models and datasets in a limited number of languages. Although the consistency of our findings across both open-source and proprietary models suggests the generalizability of our conclusions, future studies could extend this analysis to a broader range of models and across a larger group of languages. Additionally, as we do not have access to the pre-training datasets of these LLMs—and therefore cannot identify which knowledge is entirely new—we relied on fictional new knowledge. Collaboration with model developers in the future could help identify real-world examples of new knowledge for testing. Second, our study does not deeply explore the mechanisms behind these multilingual disparities. Recent research suggests the existence of language-specific and language-agnostic neurons in LLMs\cite{tang2024language}. The inequalities revealed in this study may stem from the under-training of neurons corresponding to low-resource languages. Future interdisciplinary collaborations among linguists, neuroscientists, and AI researchers could provide deeper insights into these mechanisms. Third, we have not yet proposed effective solutions to eliminate these inequalities. These disparities may originate from the biased distribution of data across different languages within pre-training datasets\cite{xie2024social}. Existing approaches, such as multilingual instruction tuning\cite{zhang2023bayling,zhu2023extrapolating,ustun2024aya} or continual pre-training\cite{cui2023efficient,kuulmets2024teaching,jaavid2024romansetu} with external high-quality multilingual datasets, can provide certain improvements but face limitations in scalability and effectiveness. Future research should explore more efficient strategies to enhance multilingual capabilities and address these disparities.

\section*{Ethical and Societal Impact}

The objective of this study is to uncover inequalities inherent in the dynamic process of learning new knowledge by LLMs across different languages. Through extensive experiments, we highlight potential harms faced by users of low-resource languages, who may struggle to obtain accurate new knowledge from LLMs and are more likely to encounter lower quality or misleading outputs. Furthermore, low-resource languages may be further marginalized in an increasingly AI-driven world. Additionally, this work aims to inspire more comprehensive research into the multilingual capabilities of LLMs and inequalities that arise among different languages in the future. It encourages the exploration of the mechanisms underlying their multilingual abilities and the development of strategies to enhance these capabilities and mitigate existing inequalities. Ultimately, the goal is to contribute to the creation of a next-generation AI that is fair, inclusive, and responsible.

\clearpage

%%%%%%%%%%%%%%%% MAIN TEXT FIGURES %%%%%%%%%%%%%%%

\begin{figure}
\centering
\includegraphics[width=\linewidth]{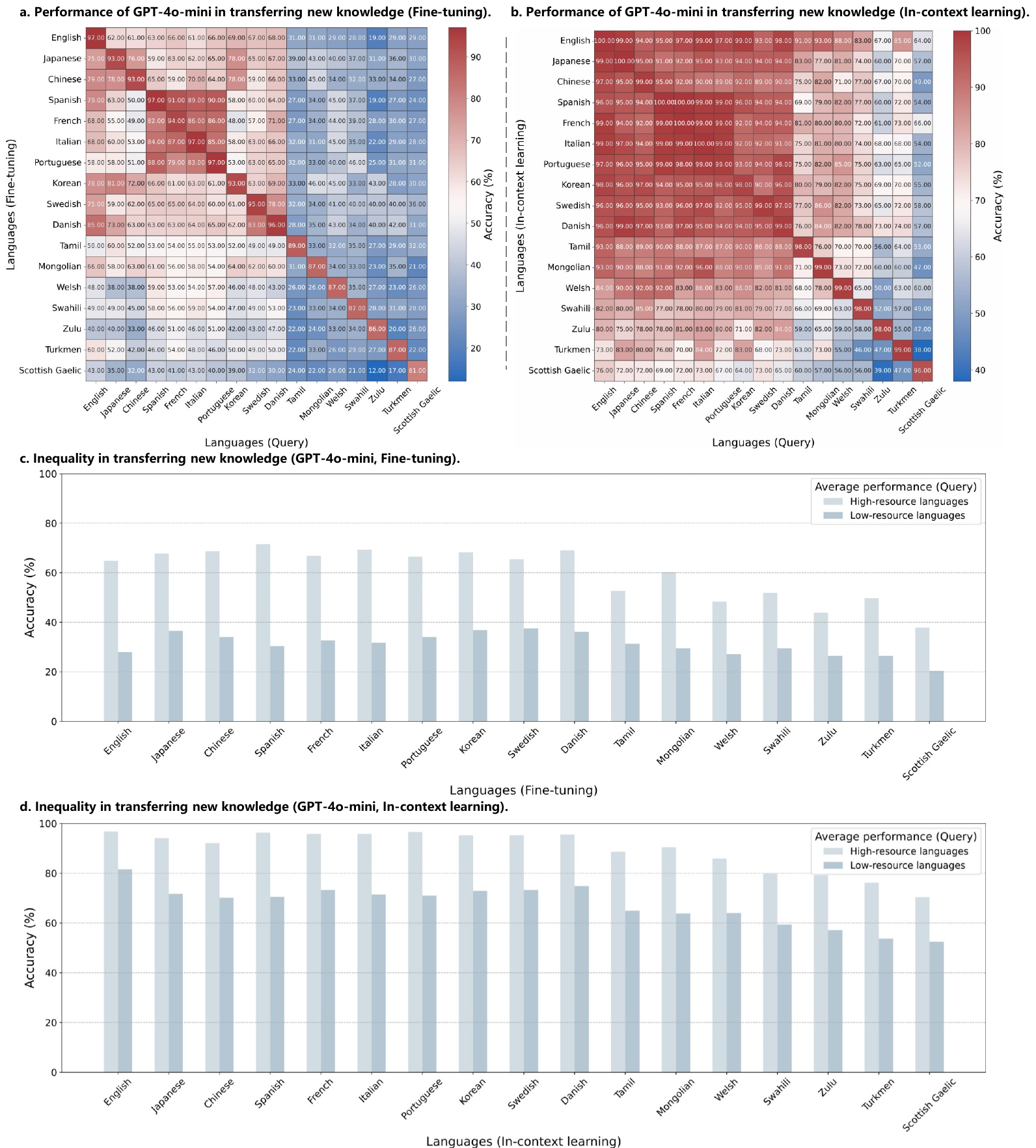}
\caption{\small \textbf{(a, c)}~The performance of GPT-4o-mini in transferring new knowledge under the fine-tuning setting and the underlying inequality. \textbf{(b, d)}~The performance of GPT-4o-mini in transferring new knowledge under the in-context learning setting and the underlying inequality. New knowledge acquired by LLMs can be more easily transferred to high-resource languages than to low-resource languages.}
\label{fig:transferable}
\end{figure}

%%%%%%%%%%%%%%%% MAIN TEXT TABLES %%%%%%%%%%%%%%%

\begin{table}
\centering
\caption{\small The 17 languages selected for our study.}
\label{tab:language}
    \setlength{\tabcolsep}{9pt}
    \begin{tabular}{lllcc}
    \hline
    \rowcolor{mygray}
    \textbf{Language} & \textbf{Family} & \textbf{Branch} & \textbf{Proportion\cite{common_crawl} (\%)} & \textbf{Resource} \\
    \hline
    English           & Indo-European   & Germanic        & 43.4241                                     & High \\
    Japanese          & Japonic         & Japanesic       & 5.0419                                      & High \\
    Chinese           & Sino-Tibetan    & Sinitic         & 4.8129                                      & High \\
    Spanish           & Indo-European   & Italic          & 4.5387                                      & High \\
    French            & Indo-European   & Italic          & 4.3960                                      & High \\
    Italian           & Indo-European   & Italic          & 2.5282                                      & High \\
    Portuguese        & Indo-European   & Italic          & 2.3146                                      & High \\
    Korean            & Koreanic        & Korean          & 0.7388                                      & High \\
    Swedish           & Indo-European   & Germanic        & 0.6649                                      & High \\
    Danish            & Indo-European   & Germanic        & 0.4640                                      & High \\
    \hline
    \rowcolor{mygray}
    Tamil             & Dravidian       & South Dravidian & 0.0473                                      & Low \\
    \rowcolor{mygray}
    Mongolian         & Mongolic-Khitan & Mongolic        & 0.0161                                      & Low \\
    \rowcolor{mygray}
    Welsh             & Indo-European   & Celtic          & 0.0117                                      & Low \\
    \rowcolor{mygray}
    Swahili           & Atlantic-Congo  & Benue-Congo     & 0.0096                                      & Low \\
    \rowcolor{mygray}
    Zulu              & Atlantic-Congo  & Benue-Congo     & 0.0025                                      & Low \\
    \rowcolor{mygray}
    Turkmen           & Turkic          & Common Turkic    & 0.0021                                      & Low \\
    \rowcolor{mygray}
    Scottish Gaelic   & Indo-European   & Celtic          & 0.0014                                      & Low \\
    \hline
    \end{tabular}
\end{table}

%%%%%%%%%%%%%%%% REFERENCES %%%%%%%%%%%%%%%

\clearpage
\bibliography{science_template}
\bibliographystyle{sciencemag}

%%%%%%%%%%%%%%%% ACKNOWLEDGEMENTS %%%%%%%%%%%%%%%

\section*{Acknowledgments}
The authors would like to express their sincere thanks to the editor and all potential anonymous reviewers for their comments and suggestions that may enhance the quality of this paper.

\paragraph*{Author contributions:}
C.W. conceived the idea of this work, conducted the experiments, analyzed the results, and contributed to the writing of this manuscript. H.T. analyzed the results. X.Y. analyzed the results. Y.X. contributed to the writing of this manuscript. J.S. contributed to the writing of this manuscript. S.S. contributed to the writing of this manuscript. J.H. contributed to the writing of this manuscript. Y.X. contributed to the writing of this manuscript. Z.G. contributed to the writing of this manuscript, and coordinated the research project. X.X. contributed to the writing of this manuscript, and coordinated the research project. F.W. conceived the idea of this work, analyzed the results, contributed to the writing of this manuscript, and coordinated the research project.

\paragraph*{Competing interests:}
There are no competing interests to declare.

\paragraph*{Data and materials availability:}
The datasets and codes applied in the experiments are publicly available at \href{https://github.com/microsoft/LNewKnow}{https://github.com/microsoft/LNewKnow}.

%%%%%%%%%%%%%%%% SUPPLEMENT LIST %%%%%%%%%%%%%%%

\subsection*{Supplementary materials}
Related Work\\
Implementation Details\\
Additional Results for Llama-3.1-8B\\
Quality Assessment of the Multilingual Parallel Datasets\\
Figure S1\\
Tables S1 to S4\\
Multilingual Parallel Fictional New Knowledge Dataset\\
Multilingual Parallel Common-sense Dataset

% %%%%%%%%%%%%%%%% END OF MAIN TEXT %%%%%%%%%%%%%%%

\newpage

% %%%%%%%%%%%%%%%% START OF SUPPLEMENT %%%%%%%%%%%%%%%

\renewcommand{\thefigure}{S\arabic{figure}}
\renewcommand{\thetable}{S\arabic{table}}
\renewcommand{\theequation}{S\arabic{equation}}
\renewcommand{\thepage}{S\arabic{page}}
\setcounter{figure}{0}
\setcounter{table}{0}
\setcounter{equation}{0}
\setcounter{page}{1}

% %%%%%%%%%%%%%%%% SUPPLEMENT TITLE PAGE %%%%%%%%%%%%%%%

\newpage
\clearpage
\appendix
\renewcommand{\thesection}{SI \arabic{section}}
\renewcommand{\thesection}{SI \arabic{section}}
\renewcommand{\thefigure}{S\arabic{figure}}
\renewcommand{\thetable}{S\arabic{table}}

\setcounter{section}{0}
\setcounter{figure}{0}
\setcounter{table}{0}

\begin{center}
\section*{Supplementary Materials for\\ \scititle}
Chenglong Wang,
Haoyu Tang$^{+}$,
Xiyuan Yang,
Yueqi Xie$^{\ast}$,
Jina Suh,
Sunayana Sitaram,\\
Junming Huang,
Yu Xie,
Zhaoya Gong$^{\ast}$,
Xing Xie,
Fangzhao Wu$^{\ast}$\\
\small$^{+}$Contribution during internship at Microsoft Research Asia\\
\small$^\ast$yxieay@connect.ust.hk, z.gong@pku.edu.cn, fangzwu@microsoft.com.
\end{center}

\subsubsection*{This PDF file includes:}
Related Work\\
Implementation Details\\
Additional Results for Llama-3.1-8B\\
Quality Assessment of the Multilingual Parallel Datasets\\
Figure S1\\
Tables S1 to S4\\
Multilingual Parallel Fictional New Knowledge Dataset\\
Multilingual Parallel Common-sense Dataset
\newpage

% %%%%%%%%%%%%%%%% MATERIALS AND METHODS %%%%%%%%%%%%%%%

\subsection*{Related Work}

\textbf{Multilingual capability of LLMs:} Most state-of-the-art LLMs, such as the Llama-3 series\cite{meta2024llama3}, are trained on multilingual datasets and can understand and generate text in multiple languages. However, the uneven distribution of data in different languages within the pre-training corpora leads to varying performance across languages. Existing research has primarily focused on building multilingual benchmarks—ranging from general NLP tasks\cite{clark2020tydi,lewis-etal-2020-mlqa,zhang2023m3exam,wang2024seaeval,zhang2024p,huang2025benchmax} to domain-specific applications\cite{qiu2024towards,niklaus-etal-2023-lextreme}—to assess the performance of LLMs across different languages. Researchers have not only collected language-specific questions to examine the amount of factual knowledge encoded in various languages but have also constructed multilingual parallel datasets to study cross-lingual consistency—i.e., the degree to which an LLM provides consistent answers to the same question posed in various languages. Findings indicate that knowledge available in low-resource languages is limited due to the lack of pre-training data in these languages\cite{jiang2020x,kassner-etal-2021-multilingual,myung2025blend}. Additionally, cross-lingual consistency is relatively low across a range of LLMs, and while larger models demonstrate improved factual accuracy, this does not necessarily enhance cross-lingual knowledge consistency\cite{qi-etal-2023-cross,chua2024crosslingual}.

\textbf{Multilingual enhancement for LLMs:} Existing methods to enhance the multilingual capabilities of LLMs or reduce language disparities can be broadly categorized into two main approaches: in-context learning-based methods and post-training-based methods \cite{zhao2024lens}. The former utilizes the inherent language translation capabilities of LLMs to improve performance. These approaches typically translate questions in low-resource languages into high-resource languages before generating responses, thereby enhancing accuracy\cite{qin2023cross,huang2023not,zhang2024autocap}. For instance, one technique known as \textit{cross-lingual thought prompting} has been proposed. This method guides LLMs to produce logical responses by following a structured process that includes problem understanding, cross-lingual reasoning, task analysis, task execution, and output formatting\cite{huang2023not}. The latter, on the other hand, focuses on modifying the model itself and can be further divided into continual pre-training\cite{cui2023efficient,kuulmets2024teaching,jaavid2024romansetu} and instruction tuning\cite{zhang2023bayling,zhu2023extrapolating,ustun2024aya}. For example, researchers have extended Llama’s vocabulary and conducted secondary pre-training using Chinese-language datasets. This process was followed by fine-tuning with Chinese-language instruction datasets, significantly improving the model’s performance in the Chinese language\cite{cui2023efficient}. Another notable approach involves creating interactive translation pairs and employing instruction tuning to transfer language generation and instruction-following capabilities from English to other languages\cite{zhang2023bayling}.

\textbf{Underlying mechanisms by which LLMs process multilingual texts:} Few studies have attempted to delve into the internal workings of LLMs in multilingual settings\cite{tang2024language,zhao2024large,kojima2024multilingual}. These studies often draw inspiration from neurobiology, which suggests that while certain brain regions involved in processing different languages may overlap, notable differences also exist\cite{friederici2011brain,parr2022active,khanna2024single}. For instance, some researchers propose that regions within LLMs can be categorized as either language-agnostic or language-specific. Language-agnostic regions are responsible for handling pragmatic principles and universal knowledge, whereas language-specific regions focus on processing language-specific vocabulary, grammar, and idiomatic expressions. To identify these regions, a method called \textit{language activation probability entropy} has been introduced. 
Experimental findings suggest that neurons related to specific languages are primarily located in the top and bottom layers of LLMs. These neurons significantly influence the model’s proficiency in processing particular languages\cite{tang2024language}.

While these studies primarily focus on exploring multilingual disparities based on the existing knowledge and capabilities of LLMs, our research shifts the focus to another critical dynamic process: learning new knowledge. Investigating inequalities in this dynamic learning process is essential for achieving a deeper and more comprehensive understanding of multilingual disparities in AI systems. These insights can further contribute to the development of next-generation AI systems that are more responsible and inclusive.

\subsection*{Implementation Details}

For the fine-tuning of GPT-4o-mini, we employed the official fine-tuning API with a batch size of 1 and a learning rate multiplier of 1.8\cite{openai_fine_tuning}. For the fine-tuning of Llama-3.1-8B, we utilized Low-Rank Adaptation (LoRA), a parameter-efficient fine-tuning technique, with a learning rate of 0.001\cite{hu2021lora}. The LoRA rank was fixed at 32, and the scaling factor was also set to 32. To evaluate the model responses, we used GPT-4o-mini. The specific prompt is provided below, and after a thorough manual check, we confirmed that GPT-4o-mini is capable of making the correct judgments.

\begin{tcolorbox}[title = {Prompt used to evaluate the model responses.}]
Given a question, a model-generated answer, and a reference answer, compare the model-generated answer with the reference answer and determine whether the generated answer is correct.

Question: \verb|question|

Generated Answer: \verb|model-generated answer|

Reference Answer: \verb|reference answer|

Output the result in the following format:

Correct: \verb|Yes/No|

Ensure that the judgment is based on semantic alignment with the reference answer.

\end{tcolorbox}

\subsection*{Additional Results for Llama-3.1-8B}

\begin{figure}
\centering
\includegraphics[width=\linewidth]{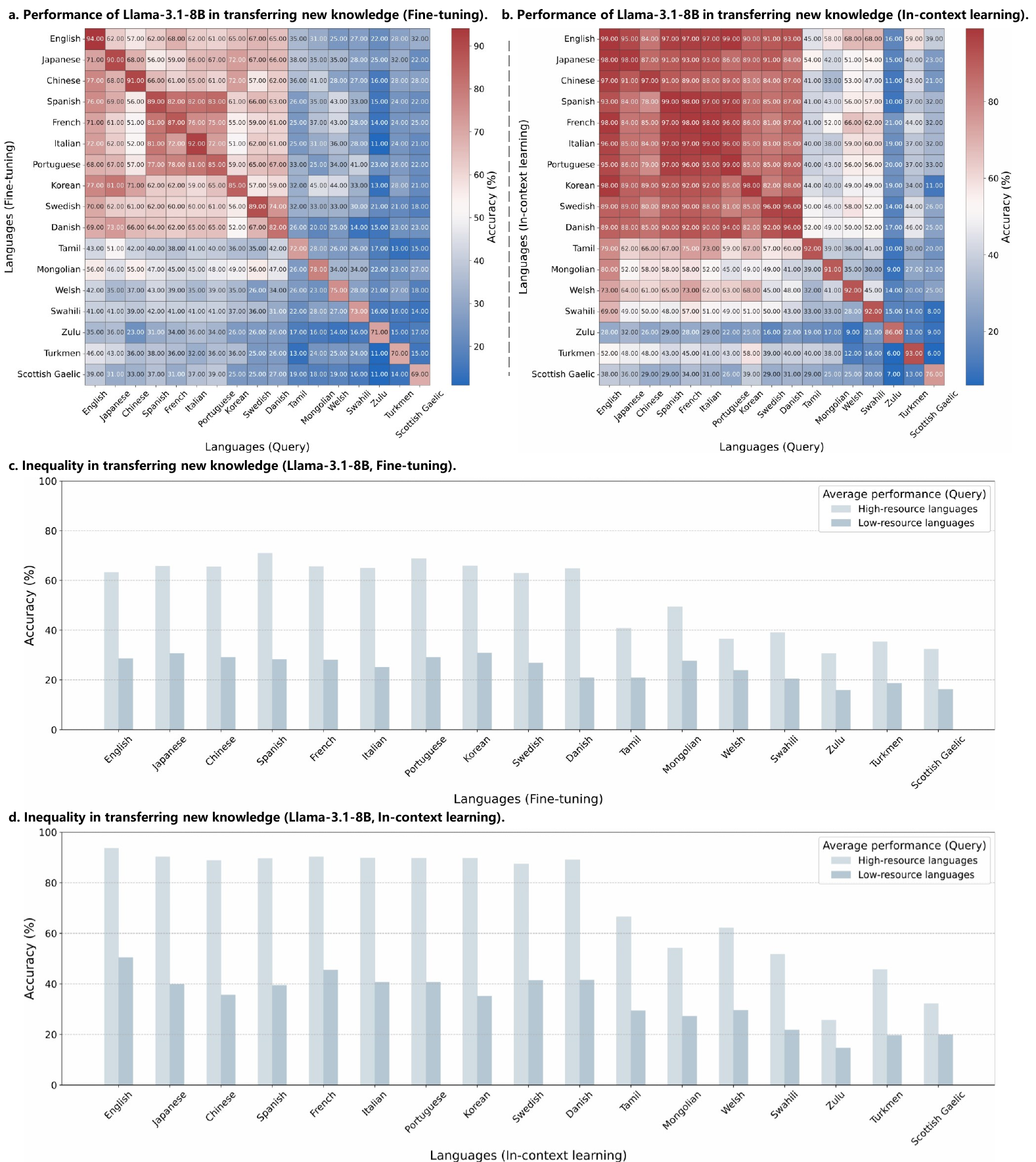}
\caption{\small \textbf{(a, c)}~The performance of Llama-3.1-8B in transferring new knowledge under the fine-tuning setting and the underlying inequality. \textbf{(b, d)}~The performance of Llama-3.1-8B in transferring new knowledge under the in-context learning setting and the underlying inequality. New knowledge acquired by LLMs can be more easily transferred to high-resource languages than to low-resource languages.}
\label{fig:transferable_sup}
\end{figure}

As shown in Figure \ref{fig:transferable_sup}, knowledge learned in one language is not always fully transferable to others, and newly acquired knowledge can be transferred more easily to high-resource languages than to low-resource ones. These findings are consistent with our experiments on the proprietary model GPT-4o-mini, highlighting the pervasive nature of these inequalities in multilingual knowledge learning and transfer.

\subsection*{Quality Assessment of the Multilingual Parallel Datasets}

We generated question–answer pairs in English and translated them into the 16 other languages using Google Translate. To assess the quality of our two multilingual parallel datasets, we conducted semantic similarity and consistency checks. This involved back-translating the question–answer pairs from the target languages into English and comparing them with the original English pairs. For semantic similarity measurement, we employed the all-MiniLM-L6-v2 model\footnote{https://huggingface.co/sentence-transformers/all-MiniLM-L6-v2} to calculate the cosine similarity between the question–answer pairs. To assess consistency, we leveraged GPT-4o-mini to evaluate whether the topics discussed and the answers provided were consistent. As illustrated in Tables \ref{table:quality_new_knowledge} and \ref{table:quality_common-sense}, both datasets exhibit high quality.

Also, as shown in Tables \ref{tab:new_knowledge_acc} and \ref{tab:common_sense_acc}, our test models struggled to accurately answer questions in any language when evaluated on the fictional new knowledge dataset. In contrast, the models successfully answered the majority of questions across all languages on the common-sense dataset. These findings highlight the effectiveness of the two datasets we constructed.

\clearpage

% %%%%%%%%%%%%%%%% SUPPLEMENTARY FIGURES %%%%%%%%%%%%%%%

\clearpage

% %%%%%%%%%%%%%%%% SUPPLEMENTARY TABLES %%%%%%%%%%%%%%%

\begin{table}
\centering
\caption{The quality of the multilingual parallel fictional new knowledge dataset.}
\label{table:quality_new_knowledge}
\setlength{\tabcolsep}{9.5pt}
\begin{tabular}{c|cc|c|cc}
\hline
\multirow{2}{*}{\textbf{Language}} & \multicolumn{2}{c|}{\textbf{Quality}} & \multirow{2}{*}{\textbf{Language}} & \multicolumn{2}{c}{\textbf{Quality}} \\ \cline{2-3} \cline{5-6}
                          & \textbf{Similarity} & \textbf{Consistency} &                           & \textbf{Similarity} & \textbf{Consistency} \\ \hline
Japanese                  & 97.3\%     & 100\%       & Danish                    & 98.3\%     & 100\%       \\ 
Chinese                   & 96.1\%     & 100\%       & Tamil                     & 95.1\%     & 100\%       \\ 
Spanish                   & 98.4\%     & 100\%       & Mongolian                 & 92.8\%     & 100\%       \\ 
French                    & 98.0\%     & 100\%       & Welsh                     & 98.6\%     & 100\%       \\ 
Italian                   & 97.6\%     & 100\%       & Swahili                   & 94.3\%     & 100\%       \\
Portuguese                & 98.1\%     & 100\%       & Zulu                      & 90.6\%     & 100\%       \\
Korean                    & 97.8\%     & 100\%       & Turkmen                   & 91.7\%     & 100\%       \\
Swedish                   & 98.0\%     & 100\%       & Scottish Gaelic           & 92.6\%     & 100\%       \\ \hline
\end{tabular}
\end{table}

\begin{table}
\centering
\caption{The quality of the multilingual parallel common-sense dataset.}
\label{table:quality_common-sense}
\setlength{\tabcolsep}{9.5pt}
\begin{tabular}{c|cc|c|cc}
\hline
\multirow{2}{*}{\textbf{Language}} & \multicolumn{2}{c|}{\textbf{Quality}} & \multirow{2}{*}{\textbf{Language}} & \multicolumn{2}{c}{\textbf{Quality}} \\ \cline{2-3} \cline{5-6}
                          & \textbf{Similarity} & \textbf{Consistency} &                           & \textbf{Similarity} & \textbf{Consistency} \\ \hline
Japanese                  & 95.8\%     & 100\%       & Danish                    & 98.0\%     & 100\%       \\ 
Chinese                   & 97.3\%     & 100\%       & Tamil                     & 96.9\%     & 100\%       \\ 
Spanish                   & 95.8\%     & 100\%       & Mongolian                 & 95.2\%     & 100\%       \\ 
French                    & 97.0\%     & 100\%       & Welsh                     & 96.9\%     & 100\%       \\ 
Italian                   & 95.2\%     & 100\%       & Swahili                   & 94.1\%     & 100\%       \\
Portuguese                & 96.4\%     & 100\%       & Zulu                      & 91.6\%     & 100\%       \\
Korean                    & 97.1\%     & 100\%       & Turkmen                   & 93.8\%     & 100\%       \\
Swedish                   & 97.9\%     & 100\%       & Scottish Gaelic           & 93.9\%     & 100\%       \\ \hline
\end{tabular}
\end{table}

\begin{table}
\centering
\caption{The performance of the LLMs on the fictional new knowledge dataset across different languages.}
\label{tab:new_knowledge_acc}
\setlength{\tabcolsep}{5.5pt}
\begin{tabular}{c|cc|c|cc}
\hline
\multirow{2}{*}{\textbf{Language}} & \multicolumn{2}{c|}{\textbf{Accuracy}} & \multirow{2}{*}{\textbf{Language}} & \multicolumn{2}{c}{\textbf{Accuracy}} \\ \cline{2-3} \cline{5-6}
                          & \textbf{GPT-4o-mini} & \textbf{Llama-3.1-8B} &                  & \textbf{GPT-4o-mini} & \textbf{Llama-3.1-8B} \\ \hline
English                   & 0.00\%               & 0.00\%                & Danish           & 1.00\%               & 0.00\% \\
Japanese                  & 1.00\%               & 0.00\%                & Tamil            & 1.00\%               & 0.00\% \\
Chinese                   & 1.00\%               & 0.00\%                & Mongolian        & 0.00\%               & 0.00\% \\
Spanish                   & 0.00\%               & 0.00\%                & Welsh            & 1.00\%               & 1.00\% \\
French                    & 0.00\%               & 0.00\%                & Swahili          & 0.00\%               & 1.00\% \\
Italian                   & 1.00\%               & 1.00\%                & Zulu             & 0.00\%               & 0.00\% \\
Portuguese                & 2.00\%               & 1.00\%                & Turkmen          & 1.00\%               & 1.00\% \\
Korean                    & 0.00\%               & 1.00\%                & Scottish Gaelic  & 0.00\%               & 0.00\% \\
Swedish                   & 0.00\%               & 0.00\%                & Average          & 0.53\%               & 0.35\% \\ \hline
\end{tabular}
\end{table}

\begin{table}
\centering
\caption{The performance of the LLMs on the common-sense dataset across different languages.}
\label{tab:common_sense_acc}
\setlength{\tabcolsep}{5.5pt}
\begin{tabular}{c|cc|c|cc}
\hline
\multirow{2}{*}{\textbf{Language}} & \multicolumn{2}{c|}{\textbf{Accuracy}} & \multirow{2}{*}{\textbf{Language}} & \multicolumn{2}{c}{\textbf{Accuracy}} \\ \cline{2-3} \cline{5-6}
                          & \textbf{GPT-4o-mini} & \textbf{Llama-3.1-8B} &                  & \textbf{GPT-4o-mini} & \textbf{Llama-3.1-8B} \\ \hline
English                   & 100.00\%             & 100.00\%              & Danish                    & 98.00\%     & 96.00\% \\
Japanese                  & 96.00\%              & 96.00\%               & Tamil                     & 86.00\%     & 72.00\% \\
Chinese                   & 100.00\%             & 100.00\%              & Mongolian                 & 84.00\%     & 62.00\% \\
Spanish                   & 100.00\%             & 98.00\%               & Welsh                     & 88.00\%     & 62.00\% \\
French                    & 100.00\%             & 100.00\%              & Swahili                   & 86.00\%     & 68.00\% \\
Italian                   & 98.00\%              & 96.00\%               & Zulu                      & 84.00\%     & 54.00\% \\
Portuguese                & 98.00\%              & 94.00\%               & Turkmen                   & 84.00\%     & 62.00\% \\
Korean                    & 92.00\%              & 90.00\%               & Scottish Gaelic           & 84.00\%     & 54.00\% \\
Swedish                   & 92.00\%              & 92.00\%               & Average                   & 92.35\%     & 82.12\% \\ \hline
\end{tabular}
\end{table}

\clearpage
\subsection*{Multilingual Parallel Fictional New Knowledge Dataset}

\begin{scriptsize}
\setlength{\tabcolsep}{0.5pt}
\begin{longtable}{clll}
\hline
\textbf{ID} & \textbf{Question} & \textbf{Answer} & \textbf{Conflicting Answer} \\
\hline
\endfirsthead

\hline
\textbf{ID} & \textbf{Question} & \textbf{Answer} & \textbf{Conflicting Answer} \\
\hline
\endhead

\hline
\endfoot

\hline
\endlastfoot

1 & What system manages traffic flow in smart cities by 2043? & AI algorithms & Smart traffic management system \\
2 & What device enhances everyday communication in 2038? & Augmented reality glasses & Augmented reality headsets \\
3 & What technology aids in language translation in 2040? & Instant translators & AI translation devices \\
4 & How do schools teach coding in 2038? & Interactive games & Coding platforms \\
5 & What tradition is revived by communities in 2050? & Local storytelling & Sustainable local farming practices \\
6 & What innovation transforms personal mobility in 2045? & Smart sidewalks & Autonomous electric vehicles \\
7 & In which year did renewable energy surpass fossil fuels? & 2030 & 2040 \\
8 & What device helps in environmental monitoring in 2050? & Smart sensors & Drone-based ecological monitors \\
9 & What new sport gains popularity by 2035? & E-sports tournaments & Drone racing \\
10 & What technique is commonly used in construction by 2042? & Modular building & 3D printing \\
11 & What type of art becomes mainstream by 2043? & Digital art & Interactive art \\
12 & How do individuals personalize their living spaces in 2050? & Smart customization & Apps \\
13 & How do people celebrate traditional festivals by 2035? & Virtual gatherings & Offline reunion \\
14 & What technology is used for food production in 2035? & Vertical farms & Lab-grown meat technology \\
15 & How do people travel long distances in 2045? & Teleportation & Supersonic air travel \\
16 & What trend in home automation becomes common by 2043? & Voice-controlled systems & AI-powered home assistants \\
17 & When did global access to clean water become universal? & 2040 & 2030 \\
18 & When did people start using brain-computer interfaces regularly? & 2043 & 2035 \\
19 & What is the preferred method for home security in 2045? & Smart surveillance & Biometric security systems \\
20 & When did education become fully personalized? & 2043 & 2040 \\
21 & How do individuals track their health in 2048? & Genetic analytics & Wearable health monitors \\
22 & How do individuals protect their digital identities in 2050? & Blockchain verification & Decentralized identity systems \\
23 & What year marked the beginning of universal basic income? & 2038 & 2030 \\
24 & What is the most popular sport in 2040? & Drone racing & eSports \\
25 & When did smart cities start implementing AI governance? & 2040 & 2050 \\
26 & What is the dominant fashion trend in 2035? & Smart clothing & Eco-friendly clothing \\
27 & When did global internet access reach 100\%? & 2042 & 2040 \\
28 & What year is it when flying cars are common? & 2045 & 2050 \\
29 & What sustainable fashion trend emerges by 2050? & Upcycled clothing & Circular fashion \\
30 & What is the primary mode of learning in 2045? & AI tutors & Immersive virtual learning \\
31 & In what year did cities become fully vertical? & 2032 & 2048 \\
32 & What unique form of public art becomes popular in 2045? & Interactive installations & Augmented reality installations \\
33 & How do communities foster inclusivity in 2035? & Universal design & Community engagement programs \\
34 & How do people share knowledge in 2032? & Knowledge networks & Knowledge sharing platforms \\
35 & What platform replaces traditional social media in 2043? & Virtual communities & Decentralized communication networks \\
36 & How do cities promote arts and culture in 2035? & Public art grants & Public art festivals \\
37 & When did humans first live on Mars? & 2040 & 2035 \\
38 & What year saw the widespread use of personal robots? & 2035 & 2040 \\
39 & How is air travel made more sustainable by 2050? & Electric planes & Hydrogen-powered aircraft \\
40 & What is the most common pet in 2038? & Robotic pets & Human-like cats \\
41 & What method do cities use to manage noise pollution in 2045? & Sound barriers & Sound-absorbing materials \\
42 & What is the common method for communication in 2035? & Brainwaves & Holographic communication \\
43 & What unique agricultural trend develops by 2045? & Edible landscaping & Urban vertical farming \\
44 & What innovation supports remote work in 2045? & Virtual offices & Virtual collaboration tools \\
45 & When did mental health apps become standard tools? & 2040 & 2035 \\
46 & In what year did virtual reality replace physical schools? & 2042 & 2040 \\
47 & What method is used to teach empathy in schools by 2035? & Role-playing simulations & Virtual reality \\
48 & When did autonomous vehicles become mainstream? & 2040 & 2048 \\
49 & How is food distributed in 2035? & Drones & Automated cars \\
50 & What process is used for building materials in 2030? & 3D printing & Recycled material construction \\
51 & What method is employed to promote local businesses in 2043? & Community marketplaces & Community-supported economy platforms \\
52 & When did urban green spaces become a priority in city planning? & 2040 & 2030 \\
53 & How is personal data managed in 2038? & Decentralized storage & User-controlled data portfolios \\
54 & How do people curate their news in 2038? & AI-driven feeds & Automatic news aggregators \\
55 & How do people participate in government in 2035? & Digital town halls & Digital civic platforms \\
56 & How is air quality monitored in cities by 2035? & IoT sensors & AI assistant \\
57 & What form of community engagement becomes prevalent in 2043? & Crowdsourced projects & Collaborative decision-making initiatives \\
58 & What new educational model is adopted in schools by 2040? & Project-based learning & Knowledge-based learning \\
59 & What form of entertainment becomes popular in 2050? & AI-generated films & Immersive experiences \\
60 & How are buildings powered in 2050? & Smart grids & Solar energy \\
61 & When did urban wildlife corridors become standard in city planning? & 2042 & 2040 \\
62 & What food trend is prevalent in 2048? & Lab-grown meat & Plant-based diets \\
63 & What method is used for personalized shopping experiences in 2045? & AI recommendations & AI personal shopping assistants \\
64 & How is wildlife conservation monitored in 2035? & Satellite imaging & Drones \\
65 & What cultural practice re-emerges in 2048? & Community farming & Traditional craftsmanship \\
66 & What practice becomes standard for remote work environments in 2040? & Flexible hours & Fixed hours \\
67 & What is the primary energy source in 2040? & Solar & Wind \\
68 & How do societies celebrate biodiversity by 2038? & Nature festivals & Biodiversity festivals \\
69 & What new form of communication is common by 2045? & Gesture-based language & Neural interface messaging \\
70 & When did virtual workspaces become the norm? & 2040 & 2035 \\
71 & When did urban farming become a necessity? & 2038 & 2030 \\
72 & What technique is used for personal security in 2045? & Biometric scanning & Blockchain technology \\
73 & When did global temperatures stabilize? & 2050 & 2040 \\
74 & What tool aids in climate change research by 2043? & Climate modeling software & AI simulations \\
75 & What medical advancement is common in 2043? & Gene editing & Regenerative medicine \\
76 & How do communities combat food deserts in 2035? & Mobile markets & Online services \\
77 & What new system is used for energy distribution in smart homes by 2050? & Peer-to-peer networks & Decentralized energy grids \\
78 & What social movement gains momentum by 2045? & Climate activism & Digital privacy rights \\
79 & How do cities enhance pedestrian safety in 2035? & Smart crosswalks & Footbridge \\
80 & What do citizens use for health monitoring in 2045? & Wearable sensors & Implanted health sensors \\
81 & What new agricultural practice is adopted in urban settings by 2045? & Aquaponics & Vertical hydroponics \\
82 & What community initiative gains traction in 2043? & Time banking & Local sustainability programs \\
83 & How do people purchase goods in 2050? & Augmented reality & Voice-activated shopping \\
84 & What method is commonly used to clean oceans in 2045? & Autonomous drones & Autonomous cleanup ships \\
85 & What is the global language in 2030? & English & Chinese \\
86 & What strategy is used for reducing food waste in 2040? & Smart inventory systems & AI-powered supply chains \\
87 & What method is used for urban heat management in 2045? & Green roofs & Reflective building materials \\
88 & What is the most significant cultural movement by 2045? & Minimalism & Global sustainability movement \\
89 & What is the primary way people experience entertainment in 2043? & Immersive simulations & Mixed reality experiences \\
90 & What is the standard workweek length in 2050? & Four days & Three days \\
91 & When did genetic testing become commonplace for preventative health? & 2040 & 2035 \\
92 & When did climate change reversal efforts begin? & 2038 & 2030 \\
93 & What approach is taken to preserve endangered languages in 2040? & Digital archives & Sound recording \\
94 & What innovation improves elderly care by 2043? & Robotic caregivers & Automatic pets \\
95 & When did self-sustaining communities start to emerge? & 2040 & 2035 \\
96 & What year did space tourism become affordable? & 2048 & 2045 \\
97 & How is water conservation achieved in urban areas by 2035? & Rainwater harvesting & Smart water systems \\
98 & What new role emerges for AI in journalism by 2038? & Fact-checking & Automated investigative reporting \\
99 & When did virtual reality therapy gain mainstream acceptance? & 2042 & 2032 \\
100 & What new approach is used for urban planning in 2040? & Data-driven design & Theory-driven design \\
\end{longtable}
\end{scriptsize}

\clearpage
\subsection*{Multilingual Parallel Common-sense Dataset}

\begin{footnotesize}
\setlength{\tabcolsep}{9pt}
\begin{longtable}{clll}

\hline
\textbf{ID} & \textbf{Question} & \textbf{Correct Answer} & \textbf{Incorrect Answer} \\
\hline
\endfirsthead

\hline
\textbf{ID} & \textbf{Question} & \textbf{Correct Answer} & \textbf{Incorrect Answer} \\
\hline
\endhead

\hline
\endfoot

\hline
\endlastfoot

1 & If you put water in the freezer, what will it turn into? & Ice & Steam \\
2 & What do bees produce? & Honey & Eggs \\
3 & How many legs does a dog have? & Four & Five \\
4 & What is the opposite of hot? & Cold & Wet \\
5 & What season follows winter? & Spring & Summer \\
6 & What do cows produce that people drink? & Milk & Honey \\
7 & What is the opposite of left? & Right & Up \\
8 & What season is usually the hottest? & Summer & Spring \\
9 & What planet do we live on? & Earth & Moon \\
10 & How many days are in a week? & Seven & Six \\
11 & What month comes after December? & January & March \\
12 & What season comes after fall? & Winter & Spring \\
13 & What do chickens lay? & Eggs & Fish \\
14 & What insect makes honey? & Bee & Pig \\
15 & What do people use to brush their teeth? & Toothbrush & Fork \\
16 & What is the opposite of fast? & Slow & Hot \\
17 & What does a thermometer measure? & Temperature & Humidity \\
18 & What is the opposite of wet? & Dry & Happy \\
19 & What is the opposite of happy? & Sad & Small \\
20 & What is the opposite of big? & Small & Sad \\
21 & What is the opposite of cold? & Hot & Big \\
22 & Which direction does the sun rise from? & East & West \\
23 & What comes after Tuesday? & Wednesday & Sunday \\
24 & What direction is opposite of south? & North & East \\
25 & What is the opposite of quiet? & Noisy & Happy \\
26 & What is 2 plus 2? & Four & One \\
27 & What color is blood? & Red & Black \\
28 & What do you call a large natural area with lots of trees? & Forest & Desert \\
29 & What color is the sky on a clear day? & Blue & Black \\
30 & What is the name of the star at the center of our solar system? & Sun & Earth \\
31 & What is the opposite of up? & Down & East \\
32 & What is the opposite of empty? & Full & Right \\
33 & How many hours are in a day? & Twenty-four & Twenty \\
34 & What shape has three sides? & Triangle & Square \\
35 & How many wheels does a bicycle have? & Two & Three \\
36 & What is the opposite of open? & Closed & Wet \\
37 & What do you call the color of snow? & White & Red \\
38 & How many months are in a year? & Twelve & Ten \\
39 & What is the opposite of weak? & Strong & Sad \\
40 & What do you call the day before today? & Yesterday & Tomorrow \\
41 & How many seasons are in a year? & Four & Three \\
42 & How many minutes are in an hour? & Sixty & One hundred \\
43 & How many letters are in the English alphabet? & Twenty-six & Twenty-five \\
44 & What do you call the part of your body you use to hear? & Ear & Mouth \\
45 & How many colors are in a rainbow? & Seven & Six \\
46 & What is the opposite of expensive? & Cheap & High \\
47 & What is the opposite of narrow? & Wide & Cheap \\
48 & What is the opposite of success? & Failure & Win \\
49 & How many centimeters are in a meter? & One hundred & Ten \\
50 & What is the opposite of maximum? & Minimum & Right \\
\end{longtable}
\end{footnotesize}

\end{document}